\documentclass{ecai} 

\usepackage{latexsym}
\usepackage{amssymb}
\usepackage{amsmath}
\usepackage{amsthm}
\usepackage{booktabs}
\usepackage{enumitem}
\usepackage{graphicx}
\usepackage{color}

\usepackage{times}
\usepackage{latexsym}
\usepackage{pifont}
\usepackage{xcolor}
\newcommand{\etal}{\emph{et al.}}
\usepackage{graphicx}
\usepackage{amsmath,amssymb} 
\usepackage{mathtools}
\usepackage{color}
\usepackage{booktabs}
\usepackage{algorithm}
\usepackage{algpseudocode}
\usepackage[export]{adjustbox} 
\usepackage{array,multirow}
\usepackage{url}
\usepackage{tikz}
\usepackage{tablefootnote}

\usepackage{xspace}

\newcommand{\BibTeX}{B\kern-.05em{\sc i\kern-.025em b}\kern-.08em\TeX}
\newcommand{\minisection}[1]{\vspace{0.005in} \noindent {\bf #1}}

\def\ourmethod{{\textit{GRIF-DM}}\xspace}

\begin{document}


\begin{frontmatter}


\paperid{634} 


\title{GRIF-DM: Generation of Rich Impression Fonts \\ using Diffusion Models}

\author[$\dagger$]{Lei Kang}
\author[$\ddagger$]{Fei Yang}
\author[$\dagger$]{Kai Wang}
\author[$\dagger$]{Mohamed Ali Souibgui}
\author[$\dagger$]{Lluis Gomez}
\author[$\dagger$]{Alicia Fornés}
\author[$\dagger$]{Ernest Valveny}
\author[$\dagger$]{Dimosthenis Karatzas}

\address[$\dagger$]{Computer Vision Center, Universitat Autònoma de Barcelona, Spain}
\address[$\ddagger$]{College of Computer Science, Nankai University, China}

\address{\{lkang, kwang, msouibgui, lgomez, afornes, ernest, dimos\}@cvc.uab.es\\
feiyang@nankai.edu.cn}

\begin{abstract}
Fonts are integral to creative endeavors, design processes, and artistic productions. The appropriate selection of a font can significantly enhance artwork and endow advertisements with a higher level of expressivity. Despite the availability of numerous diverse font designs online, traditional retrieval-based methods for font selection are increasingly being supplanted by generation-based approaches. These newer methods offer enhanced flexibility, catering to specific user preferences and capturing unique stylistic impressions. 
However, current impression font techniques based on Generative Adversarial Networks (GANs) necessitate the utilization of multiple auxiliary losses to provide guidance during generation. Furthermore, these methods commonly employ weighted summation for the fusion of impression-related keywords.  This leads to generic vectors with the addition of more impression keywords, ultimately lacking in detail generation capacity. 
In this paper, we introduce a diffusion-based method, termed \ourmethod, to generate fonts that vividly embody specific impressions, utilizing an input consisting of a single letter and a set of descriptive impression keywords. 
The core innovation of \ourmethod lies in the development of dual cross-attention modules, which process the characteristics of the letters and impression keywords independently but synergistically, ensuring effective integration of both types of information. 
Our experimental results, conducted on the MyFonts dataset, affirm that this method is capable of producing realistic, vibrant, and high-fidelity fonts that are closely aligned with user specifications. This confirms the potential of our approach to revolutionize font generation by accommodating a broad spectrum of user-driven design requirements. Our code is publicly available at \url{https://github.com/leitro/GRIF-DM}.
\end{abstract}

\end{frontmatter}


\section{Introduction}

Fonts constitute pivotal elements within the domain of creativity, design, and visual communication~\cite{amare2012seeing,singla2022understanding}. The judicious choice of a suitable font holds the potential to substantially augment the impact of artistic endeavors, streamline design workflows, and infuse advertisements with expressiveness.  Nowadays, there exists convenient access to an extensive array of over 270,000 fonts encompassing diverse designs, readily accessible online~\footnote{https://www.myfonts.com/}. 
Font selection traditionally relies on retrieval-based methods, wherein users sift through extensive font libraries to identify the most fitting option. However, with the continual evolution of creative demands, there emerges a necessity for more adaptable and flexible approaches to font generation. Recently, generation-based techniques have emerged as a promising alternative, providing flexibility to accommodate user preferences and manifest distinct impression concepts. An ideal font generation methodology should exhibit diversity in generating predefined impression keywords and should be capable of accommodating variable-length combinations of these keywords as conditions.

\begin{figure}[t!]
    \centering
    \includegraphics[width = 0.9999\linewidth]{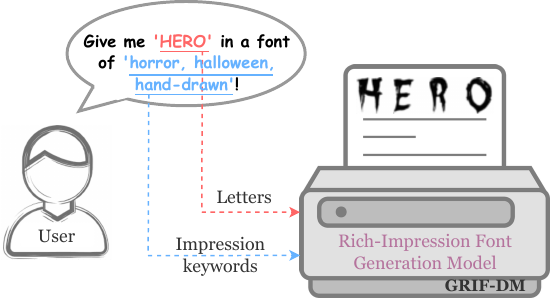}
    \caption{The illustration of the problem setup of font generation.
    Our method, \ourmethod, generates desired fonts based on user input of impression keywords and letters.}
    \label{fig:front}
\end{figure}
\vspace{0.4cm}

Font style transfer, as demonstrated in prior works~\cite{azadi2018multi,zhang2018separating,wen2021zigan,kang2021content}, has proven successful in font generation tasks. 
By furnishing textual content alongside visual style information, these models can generate synthetic fonts that emulate the specified style. Such methodologies prove particularly advantageous in situations where only a subset of characters from a font is accessible. In our scenario of rich impression font generation, users do not furnish a template font image for style information. Rather, they articulate their impressions through a list of keywords in natural language, and we anticipate the model to generate novel fonts based on this input.

Following this idea, some Generative Adversarial Networks (GANs) based methods~\cite{matsuda2021impressions2font,matsuda2022font,wang2020attribute2font} are proposed to generate new fonts based on user queries provided as attributes. 
However, these approaches all contend with the challenge of employing multiple auxiliary losses for generating target new font images. GAN-based methods, which inherently rely on a binary discriminator loss to differentiate between fake and real samples, require additional auxiliary losses such as attribute loss for precise style control and character classification loss to maintain the desired letter with fine-grained fidelity. Moreover, GAN-based methods often encounter unstable training dynamics and convergence issues due to the complexities inherent in adversarial training schemes.

More recently, Denoising Diffusion Probabilistic Models (DDPMs), also named diffusion models~\cite{ho2020denoising}, have emerged as a new family of generative models, potentially replacing GANs.
In contrast to GAN-based models that directly estimate the data distribution, diffusion models / DDPMs function by iteratively diffusing noise throughout a provided input to generate samples. This iterative process involves simulating the gradual propagation of noise across the input space, resulting in the creation of realistic samples. A notable advantage of DDPMs lies in their capacity to generate high-quality images characterized by coherent structures and fine details. This attribute renders DDPMs particularly suitable for font generation tasks, wherein the preservation of intricate details and the minimization of noise are imperative for producing visually captivating outcomes.


In this paper, we introduce \ourmethod, a novel diffusion-based method for generating rich impression fonts.
Our approach takes a single letter and a set of impression keywords as input, allowing for the customization of fonts to specific user preferences. Drawing inspiration from the DDPM~\cite{ho2020denoising}, we devise a U-Net architecture comprising encoder, bottleneck, and decoder modules. To effectively integrate letter and impression information, we introduce dual cross-attention modules. Given the variable number of impression keywords associated with each font, we concatenate them into a sentence format using commas as separators and leverage a pre-trained BERT model to extract variable-length textual embeddings. Similarly, we embed the letter using another pre-trained BERT model. Through cross-attention mechanisms, both impression and letter embeddings are seamlessly integrated with visual font features. Exploiting BERT's semantic comprehension of text, our method exhibits reasonable performance in handling out-of-vocabulary impression keywords by treating them as synonyms of keywords present in the training set.

As the summarization, the main contributions of this paper are:

\begin{itemize}
    \item To the best of our knowledge, we are the first to introduce a diffusion-based approach for impression font generation with the English alphabet, thereby eliminating the need for additional auxiliary losses observed in recent GAN-based methods.
    \item We introduce novel dual cross-attention modules designed to effectively integrate information from both letters and impression keywords.
    \item We propose merging impression keywords into sentences rather than employing weighted sums of individual impression vectors. This approach ensures robustness and preserves fine details for specific impression keywords, even when dealing with a large set of impression keywords.
\end{itemize}

\section{Related Work}

\subsection{GAN based Font Generation}

GAN-based models consist of a generator and a discriminator, which are trained simultaneously in a competitive manner. The generator learns to produce realistic samples, such as font images, while the discriminator learns to distinguish between real data and generated samples. This adversarial training process encourages the generator to continually improve its ability to produce high-quality outputs that are indistinguishable from real data. While for font generation, GAN-based methods need extra auxiliary losses to control the desired content and style effectively. 
zi2zi~\cite{tian2017zi2zi} proposed synthesizing Chinese calligraphy images, while controlling them by conditioning on a category information. Xie~\etal~\cite{xie2021dg} proposed a new deformable DGFont for unsupervised font generation. Hayashi~\etal~\cite{hayashi2019glyphgan} proposed GlyphGAN to control the font style while maintaining character style consistency. Wang\etal~\cite{wang2021deepvecfont} proposed DeepVecFont, which generates font images in a vector format. Wang~\etal~\cite{wang2020attribute2font} proposed Attribute2Font, which synthesizes a font from specific impressions. However, the number of attributes is limited to 37. Matsuda~\etal~\cite{matsuda2021impressions2font} proposed Impressions2Font, which generates fonts directly from impression keywords. Matsuda~\etal~\cite{matsuda2022font} proposed a font generation method trained while completing the missing labels as much as possible.

\begin{table}[t!]
    \caption{Fonts and their corresponding impression keywords from MyFonts dataset~\cite{chen2019large}. The example fonts are "vidok-fy", "trump-gothic-pro", "neil-bold", "college-dropout-senior", and "spanish-main" from top to bottom, respectively. Only letters "Q", "U", "I", "C", "K", "F", "O", and "X" are shown.}
    \vspace{0.4cm}
    \label{tab:data}
    \centering
    \small
    \scalebox{0.99}{
    \begin{tabular}{c}
    \toprule
    \includegraphics[width=0.6\linewidth]{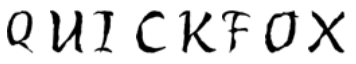}\\
    handwrite, vampire, romantic, cinema, movie, horror, \\
    invitation, texture, hand, zombie, script, 1900s, pen, \\
    retro, ink, wed, cursive, calligraphy, and 1800s.\\
    \midrule
    \includegraphics[width=0.6\linewidth]{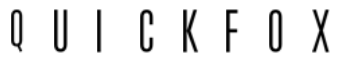}\\
    unicase, fashion, athletic, modern, monocase, economic, \\
    gothic, advertise, sport, film, cyrillic, generic, czech, \\
    grotesque, condense, german, economical, grotesk, legible, \\
    linear, poster, standard, headline, news, news-headline, \\
    condensed-gothic, realist, sturdy, movie-credits, greek, \\
    magazine, swiss, skinny, and narrow.\\
    \midrule
    \includegraphics[width=0.6\linewidth]{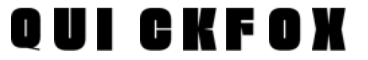}\\
    heavy, modern, 1960s, slit, signage, beefy, book-cover, \\
    unusual, bi-form, fashionable, alternate, science-fiction,\\
    black, retro, minimal, urban, bold, art-deco, ultra-bold,\\
    poster, jazz, fat, unique, sturdy, sans-serif, american,\\
    block, 1970s, wild, ultra, idiosyncratic, and nightclub.\\
    \midrule
    \includegraphics[width=0.6\linewidth]{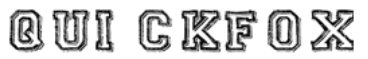}\\
    ketch, outline, hand-drawn, slab-serif, college, and chamfer.\\ 
    \midrule
    \includegraphics[width=0.6\linewidth]{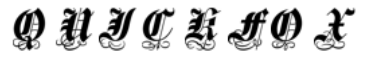}\\
    decorative, manicule, ornate, fancy, gothic, motorcycle, wed,\\
    signage, calligraphic, letterhead, 1800s, tattoo, ancient, label,\\
    certificate, headline, vintage, royal, italic, pirate, poster, \\
    english, old-english, blackletter, calligraphy, greek, fraktur, \\
    oktoberfest, revival, invitation, newspaper, xmas, and wine.\\
    \bottomrule
    \end{tabular}
    }
\end{table}

\subsection{Diffusion Model based Font Generation}

Diffusion models~\cite{ho2020denoising} represent a novel research line of generative models, showcasing their potential to surpass GAN-based methods with many successful applications in text-to-image~\cite{saharia2022photorealistic,zhang2023adding}, text-based image editing~\cite{hertz2022prompt,kai2023DPL,brooks2022instructpix2pix,tang2023iterinv,tumanyan2022plug}, object detection~\cite{chen2022diffusiondet}, image segmentation~\cite{gu2022diffusioninst,ma2023diffusionseg,xu2023odise,pnvr2023ldznet}, landmark detection~\cite{wu2024diffusion}
and more relatedly the multi-object tracking (MOT)~\cite{DiffusionTrack,lv2024diffmot,li2023ovtrack}. Diffusion models generally learn a denoising model to gradually denoise from an original common distribution, e.g. Gaussian noise, to a specific data distribution.
It performs a parameterized Markov chain to produce samples of a certain data distribution after a number of steps. In the forward direction, the Markov chain gradually adds noise to the data until it is mapped to a simple isotropic Gaussian distribution.
As a consequence of the Markov principle, DDPM~\cite{ho2020denoising} exhibits a relatively slower sampling. To address this, various sampling techniques~\cite{NEURIPS2022_260a14ac,meng2023distillation} are developed to enhance the denoising speed. 
An example is DDIM~\cite{DDIM}, which introduced a deterministic non-Markovian process to accelerate the sampling process while producing high quality generations.

More recently, Diffusion Models have also been applied in Font generation. He~\etal~\cite{he2022diff} proposed Diff-Font as the first Diffusion Model based Chinese font generation approach.
Yang~\etal~\cite{yang2024fontdiffuser} proposed FontDiffuser, which generates Chinese font images by diffusion model in the few-shot approach. Tanveer~\etal~\cite{tanveer2023ds} proposed DS-Fusion that generates the cat-like character image with the input prompt “cat.” Wang~\etal~\cite{wang2023anything} also generate artistic font consisting of “pasta” by the prompt “pasta.” These diffusion models for font generation show the great capacity to generate various decorative font images.
However, the English alphabet font generation conditioned on impressions, which is the main topic of this paper, has never been explored from the view of Diffusion Models.

\section{Method}

\subsection{Problem Formulation}
To formulate the problem of using a diffusion model to generate rich impression font images, we denote the dataset as $\mathcal{D}$. Each entry $(X_i, Y_i) \in \mathcal{D}$ represents a font category $X_i$ and its corresponding set of impression keywords $Y_i$. Here, $X_{i}^{k}$ is the $k$ letter image of $i$-th font, where $k \in \{``A", ``B", ..., ``Z"\}$. For simplicity, we represent a real font letter image $X_{i}^{k}$ as $x_0$ in the literature. Let $\theta$ represent the parameters of the diffusion model, which aims to learn the underlying distribution of font images in $\mathcal{D}$. The goal is to generate a letter image $x_0$ from a noise image $x_T \sim \mathcal{N} (0, I)$ by iteratively generating denoised images $x_{T-1}, ..., x_t, ..., x_0$. The noise $\epsilon_\theta (x_0|x_t, t, k, Y_i)$ can be estimated by taking into account the noisy image $x_t$, the current time step $t$, the letter $k$ and the given set of impression keywords $Y_i$. A few examples of $(X_i, Y_i)$ can be found in Tab.~\ref{tab:data}.

\begin{figure*}[!t]
    \centering
    \includegraphics[width = 1.01\linewidth]{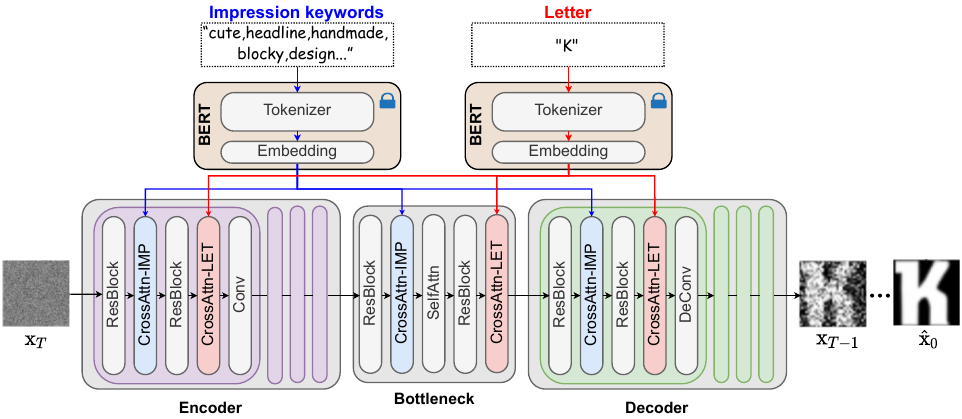}
    \caption{Architecture of our proposed method, which includes an Encoder, a Bottleneck, and a Decoder modules in purple, grey, and green respectively. Two frozen weights BERT modules are used to obtain embeddings of impression keywords and a letter, integrated via dual cross-attention modules: impression cross-attention in blue and letter cross-attention in red.
    }
    \label{fig:arch}
\end{figure*}
\vspace{0.4cm}

\subsection{U-Net}
Our proposed U-Net model adheres to the typical architecture, which consists of an encoder, a bottleneck, and a decoder module, as shown in Fig.~\ref{fig:arch}. 

\textbf{The Encoder} is composed of four repeated convolutional and linear blocks, shown in the bottom-left of Fig.~\ref{fig:arch}. It takes as input a noisy font image $x_t$, where $t \in \{T, T-1, ..., 1\}$. Each "ResBlock" contains convolutional and linear layers with residual connections, extracting high-level features while preserving input height and width while enhancing feature depth. The final “Conv” layer reduces spatial dimensions, resulting in more compact latent features $F_{enc}$.

\textbf{The Bottleneck} module comprises two “ResBlock” modules, similar to those in the Encoder. Additionally, it incorporates a self-attention module to fuse contextual information from global contexts. This integration allows the model to capture dependencies across various scales and effectively distill essential features for further processing. It takes as input the encoder feature $F_{enc}$ and produce a bottleneck feature $F_{btl}$ of same size. 

\textbf{The Decoder} comprises four iterative convolutional, linear, and deconvolutional blocks, illustrated in the bottom-right of Fig.~\ref{fig:arch}. Similar to the Encoder and Bottleneck, it employs “ResBlock” modules. However, it differentiates itself through the inclusion of “DeConv” layers, which expand input dimensions while reducing depth. This process aims to reconstruct realistic output from compressed feature representations. The Decoder takes as input 
the bottleneck feature $F_{btl}$ and produce a single-channel noise prediction $\hat{\epsilon}$ with the same size as the input of encoder $x_t$.

\subsection{Text Embedding Modules}
We employ pre-trained BERT~\cite{bert} tokenizer and text encoder modules to handle both the letter and impression keywords. The BERT tokenizer operates at the word-piece level but can also tokenize individual letters. For letter BERT, we set "max\_seq\_length" to 3, including start and end tokens, while for impression BERT, it's configured to 512, allowing for varying sequence lengths of impression keywords. Thus, we can obtain a variable length impression embedding feature $c_{imp} = BERT(Y_i)$ and a letter embedding feature $c_{let} = BERT(k)$. Utilizing pre-trained BERT embeddings allows us to capture contextual information and semantic meaning from the input text. This enables our model to seamlessly integrate both letter and impression keywords, thereby improving the comprehension and generation of lifelike font images. Leveraging pre-trained BERT modules ensures efficient text processing and robust representation learning, thereby enhancing the efficacy of our proposed framework.

\subsection{Dual Cross-attention Modules}
\label{sec:dual}
In the U-Net generation process, addressing the length discrepancy between the single-character letter input and variable-length impression keywords presents a challenge. To overcome this, we introduce a dual cross-attention module, as depicted as blue and red blocks in Fig.~\ref{fig:arch}. This module initially incorporates impression BERT feature $c_{imp}$ using cross-attention (illustrated by blue arrows labeled as “CrossAttn-IMP”) and subsequently integrates letter BERT feature $c_{let}$ using another cross-attention mechanism (indicated by red arrows labeled as “CrossAttn-LET”). We seamlessly integrate these dual cross-attention modules into the encoder, bottleneck, and decoder components. These modules operate in tandem with convolutional and linear blocks, facilitating the effective integration of both letter and impression information. Consequently, our approach remains invariant to the variable length of impression keywords, ensuring that excessively long impression keywords do not unduly impact the letter information.

\subsection{Training Process}

In this paper, we hypothesize that a real font image $X_{i}^{k}$ is determined by the letter $k$ and impression keywords $Y_i$. Simplifying, we denote the real font image $X_{i}^{k}$ as $x_0$, where $x_0 \sim q(x_0 | [c_{let}, c_{imp}])$. Here, $[c_{let}, c_{imp}]$ denotes the dual embedded latent features as conditions. We then iteratively add random Gaussian noise to $x_0$ for $T$ times, transitioning it from a stable state $x_0$ to a chaotic state $x_T$. This iterative process is termed the diffusion process and is defined as follows:

\begin{equation}
    q(x_{1:T} | x_0) = \prod_{t=1}^{T} q(x_t | x_{t-1})
\end{equation}
where each step diffusion is:
\begin{equation}
    q(x_t | x_{t-1}) = \mathcal{N}(x_t; \sqrt{1 - \beta_t}x_{t-1}, \beta_t \textbf{I}), \quad t \in \{1, \ldots, T\}
\end{equation}

$\beta$ is an ascending variance schedule from 0 to 1 following DDPM~\cite{ho2020denoising}. Using the notation $\alpha_t = 1 - \beta_t$ and $\bar{\alpha}_t = \prod_{i=1}^{t} \alpha_i$, we can obtain $x_t$ at an arbitrary timestep $t$ as the following: 

\begin{equation}
    q(x_t|x_0) = \mathcal{N}(x_t; \sqrt{\bar{\alpha}_t}x_0, (1 - \bar{\alpha}_t)\textbf{I})
\end{equation}

Thus, we can obtain $x_t$ as:

\begin{equation}
    x_t = \sqrt{\bar{\alpha}_t}x_{0} + \sqrt{1 - \bar{\alpha}_t}\epsilon, \quad \epsilon \sim \mathcal{N}(0, I)
\end{equation}

In the reverse process, our proposed model is to generate the designated font image by denoising the $x_T$ in the Markov chain by taking the dual embedded latent feature $[c_{let}, c_{imp}]$ as letter and impression condition-pair. We denote the joint distribution $p_\theta(x_{0:T} | [c_{let}, c_{imp}])$ as the reverse process Markov chain with learned Gaussian transitions starting at $p(x_T) = \mathcal{N}(x_T; \textbf{0}, \textbf{I})$. Thus,

\begin{equation}
    p_{\theta}(x_{0:T} | [c_{let}, c_{imp}]) = p(x_T) \prod_{t=1}^{T} p_{\theta}(x_{t-1}|x_t, [c_{let}, c_{imp}])
\end{equation}

Then, we can formulate the reverse step-by-step denoising as:

\begin{multline}
    p_{\theta}(x_{t-1}|x_t, [c_{let}, c_{imp}]) = \\
    \mathcal{N}(x_{t-1}; \mu_{\theta}(x_t, t, [c_{let}, c_{imp}], \Sigma_{\theta}(x_t, t, [c_{let}, c_{imp}]))
\end{multline}

Following DDPM~\cite{ho2020denoising}, we set $\Sigma_{\theta}(x_t, t, [c_{let}, c_{imp}])$ as constants and the diffusion model $\epsilon_\theta (x_t, t, [c_{let}, c_{imp}])$ learns to predict the noise $\epsilon$ added to $x_0$ in diffusion process from $x_t$ with condition $[c_{let}, c_{imp}]$. Finally, the denoising training process can be summarized as:

\begin{equation}
    L = \mathbb{E}_{x_0, \epsilon, k, Y_i} \| \epsilon - \epsilon_{\theta}(x_t, t, [c_{let}, c_{imp}]) \|^2
\end{equation}
where $x_0 \sim q(x_0)$, $\epsilon \sim \mathcal{N}(0, I)$, $k \in \{``A", ``B", ..., ``Z"\}$, and $Y_i \in Y$.

\section{Experiments}
\subsection{Dataset}
We utilize the MyFonts dataset~\cite{chen2019large} for all our experiments. Following the approach outlined in~\cite{matsuda2021impressions2font,matsuda2022font}, we select uppercase letters ranging from “A” to “Z” in the dataset. However, unlike Matsuda~\cite{matsuda2022font}, who conducted manual inspections to remove non-alphanumeric characters, we employ the following rules to automatically filter out unwanted characters: First, we discard fonts with fewer than 5 impression keywords to ensure that the remaining fonts are more specific and tailored; second, we remove fonts with a width-to-height ratio exceeding 2:1, effectively filtering out certain dingbat characters in an automated manner. Finally, we shuffle the fonts and randomly divide them into training and test sets, with a ratio of 90\% for training and 10\% for testing. The statistics of the prepared dataset are presented in Tab.~\ref{tab:stat}. Note that the number of fonts in the third column multiplied by 26 letters equals the number of images in the second column. The number of impression keywords in the fourth column represents the unique ones in each set, while the minimum, average, and maximum number of keywords per font are counted for each font.

\begin{table}[t!]
    \caption{Statistics of both training and test sets. “Imp. K.” represents Impression Keywords.}
    \vspace{0.4cm}
    \label{tab:stat}
    \centering
    \small
    \scalebox{0.99}{
    \begin{tabular}{ccccccc}
    \toprule
    \multirow{2}{*}{\textbf{Set}} & \multirow{2}{*}{\textbf{Images}} & \multirow{2}{*}{\textbf{Fonts}} & \multirow{2}{*}{\textbf{Imp. K.}} & \multicolumn{3}{c}{\textbf{Imp. K. / Font}} \\
    & & & &  Min. & Avg. & Max. \\
    \midrule
    Train Set & 347,724 & 13,374 & 1,823 & 5 & 18.4 & 184\\
    Test Set & 38,610 & 1,485 & 1,658 & 5 & 17.7 & 176\\
    \bottomrule
    \end{tabular}
    }
\end{table}

\subsection{Implementation Details}
Our implementation of the diffusion framework is built from scratch, drawing inspiration from DDPM~\cite{ho2020denoising} and DDIM~\cite{song2020denoising}. We utilize a batch size of 256 and a learning rate of $2 \times 10^{-4}$ with a step scheduler that decreases by 90\% every 10 epochs. Training for the DDPM model is conducted with T set to 1,000 time steps, while evaluation for the DDIM model is performed with T set to 100 time steps to enhance evaluation speed. Text Embedding Modules employ BERT with pre-trained weights from "google-bert/bert-base-uncased", while font images are pre-processed into grayscale and resized to 32x32 pixels. The model is trained on the MyFonts training set for 400 epochs using a single NVIDIA A40 GPU, with an Adam optimization algorithm. More details can be found in our code.

\subsection{Quantitative Results}
We utilize FID~\cite{heusel2017gans} and Intra-FID~\cite{miyato2018cgans} for the quantitative evaluation. FID measures the diversity and quality of generated font images with the pre-trained Inception Neural Network. Following the same amount of 26 $\times$ 5,000 generated samples in~\cite{matsuda2022font}, we randomly select 5,000 fonts from the whole 14,859 fonts of the dataset (13,374 fonts from training set and 1,485 fonts from test set) as the groundtruth, and make use of all the impression keywords to each font to generate synthetic font images. This setup provides a fair comparison with other methods in Tab.~\ref{tab:fid} as it utilizes the same number of randomly selected 5,000 fonts. Our proposed \ourmethod has achieved an FID of 6.693, outperforming other GAN-based methods. Additionally, in a more rigorous assessment, we compute FID using the entire test set of 1,485 unseen fonts. \ourmethod achieves an FID of 8.347, slightly inferior to the 5,000 random samples scenario, yet still outperforms other GAN-based methods.

For Intra-FID, we adopt the approach outlined in~\cite{matsuda2022font}, selecting only frequent impression keywords associated with over 200 fonts to ensure sufficient samples per class. This results in 277 impression keywords from all the dataset (training and test set). For each keyword, we generate 5,200 synthetic images (200 fonts multiplied by 26 letters). As fonts are grouped with their respective impression keywords, synthetic font images are generated based solely on the specified impression keyword as condition. This presents a challenging scenario where real fonts encompass diverse styles, yet the generated font images are restricted to a single impression keyword for generation. It evaluates the diversity capacity under the constraint of a single impression keyword. Intra-FID is calculated as the average FID across all impression classes. \ourmethod achieves an Intra-FID of 43.119, surpassing other GAN-based methods. Nevertheless, there is still room for improvement in enhancing the diversity capacity of \ourmethod when provided with a single impression keyword.

\begin{table}[t!]
    \caption{Quantitative evaluation results in FID and Intra-FID, with lower values indicating better performance. Due to the limited number of fonts per impression keyword in the test set (1,485), Intra-FID calculation for GRIF-DM (1,485) is not feasible. }
    \vspace{0.4cm}
    \label{tab:fid}
    \centering
    \small
    \resizebox{0.83\columnwidth}{!}{
    \begin{tabular}{ccc}
    \toprule
    \textbf{Method} & \textbf{FID}$\downarrow$ & \textbf{Intra-FID}$\downarrow$\\
    \midrule
    C-GAN+~\cite{mirza2014conditional} & 29.618 & 52.199\\
    AC-GAN+~\cite{odena2017conditional} & 29.152 & 68.355\\
    CP-GAN+~\cite{kaneko2018class} & 30.412 & 152.398\\
    Imp2Font~\cite{matsuda2021impressions2font} & 24.543 & 146.691\\
    FontGen~\cite{matsuda2022font} & 21.895 & 56.733\\
    \midrule
    \textbf{GRIF-DM (5,000)} & \textbf{6.693} & \textbf{43.119} \\
    \textbf{GRIF-DM (1,485)}\tablefootnote{This evaluates performance on the unseen fonts in the full test set, stricter than other methods but still demonstrates superior FID performance.} & \textbf{8.347} & $-$\\
    \bottomrule
    \end{tabular}
    }
\end{table}

\subsection{Font Diversity}
\begin{figure}
    \centering
    \includegraphics[width = 1.033\linewidth]{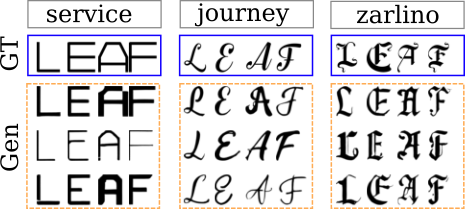}
    \caption{Qualitative results for font diversity. The top row displays three font names from the test set. Groundtruth images of "L," "E," "A," and "F" are enclosed in blue boxes, while generated font images are in orange dashed boxes, with each row starting from different random noise.}
    \label{fig:diversity}
\end{figure}

To qualitatively assess font diversity in our \ourmethod, we randomly select three unseen fonts from the test set, as indicated by the blue boxes with font names at the top of Fig.~\ref{fig:diversity}, given different random noise at $x_T$, we can generate different font images as shown in the orange dashed boxes. 

For the font "service", the impression keywords are quare, dance, bitmap, fashion, 2000s, flyer, sans-serif, monoline, polygonal, music, avant-garde, 1990s, line, future, wide, rectangular, magazine, techno, and futuristic. All three types of generated outputs exhibit coherence but display variations in styles, such as thickness and the slant of the top of the letter "A". 

For the font "journey", the impression keywords are fashion, 1950s, package, vintage, modern, ornament, fancy, casual, brush, letter, smooth, swash, and sign-painting. The generated samples effectively convey the desired impressions, showcasing diversity in elements like the left or right lean of the letter "A" and the degree of cursiveness in letters "L" and "F". 

The last font "zarlino" has impression keywords: decorative, calligraphy, fraktur, elegant, blackletter, ornate, tatoo, and tendril. In this challenging scenario, where font images contain intricate details, especially at the end of strokes, the generated font images successfully convey the intended impressions and exhibit a high fidelity to the real images. The first set of generated font images adopts a simplified style compared to the real images. Conversely, the third set closely mimics the real ones, maintaining the thin stroke style at the end. Although the letter "E" in the second set diverges from the observed cursive stroke in the real image, it adeptly conveys the intended impression keywords. 

Therefore, based on the qualitative findings, it is evident that our proposed model \ourmethod can generate diverse font images while effectively preserving the desired impression information by utilizing different random noise as a starting point.

\vspace{0.4cm}

\subsection{Exploration on Impression Keywords}

\begin{figure}
    \centering
    \includegraphics[width = 1.02\linewidth]{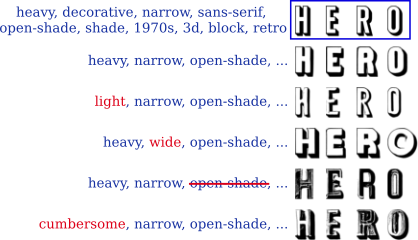}
    \caption{Exploration of Impression Keywords. Groundtruth font images of letters "H", "E", "R", and "O" enclosed in blue boxes with corresponding impression keywords to the left in the first row. Subsequent font images are generated by \ourmethod conditioned on the corresponding impression keywords to the left, with modifications highlighted in red.}
    \label{fig:ood}
\end{figure}

We conduct an exploration experiment for impression keywords, depicted in Fig.~\ref{fig:ood}. The first row features a real font from the test set, displaying letters "H", "E", "R", and "O", along with the full impression keywords listed on the left. In the second row, generated samples using the same full impression keywords as input are showcased. To clarify our concept, we emphasize the three primary impression keywords "heavy", "narrow" and "open-shade" while excluding the others, yet we utilize all impression keywords for generation. Yet as shown in the second row, the letters appear slightly thicker than the ground truth due to the semantic contrast between the impression keyword "heavy" and "narrow". To validate our intuition, we replace "heavy" with its antonym "light" while maintaining the remaining impression keywords unchanged, resulting in notably thinner generated font images in the third row. Conversely, replacing the impression keyword "narrow" with its antonym "wide" while keeping the remaining impression keywords unchanged yields wider font images in the fourth row. 

Additionally, removing the impression keyword "open-shade" darkens the generated font images in the fifth row while still conveying the intended impression. Lastly, in the final row, we replace the impression keyword "heavy" with its synonym "cumbersome". It is evident that the generated font images effectively convey the intended impression keywords, despite "cumbersome" not being present in the dataset. Generated font images make sense because leveraging BERT brings synonymous keywords closer in the feature space. By employing impression sentences with cross-attention instead of strict impression vectors, our method demonstrates robustness to out-of-vocabulary (OOV) impression keywords.

\vspace{0.4cm}

\subsection{Qualitative Comparison with SoTA}

\begin{figure*}[t!]
    \centering
    \includegraphics[width = 0.98\linewidth]{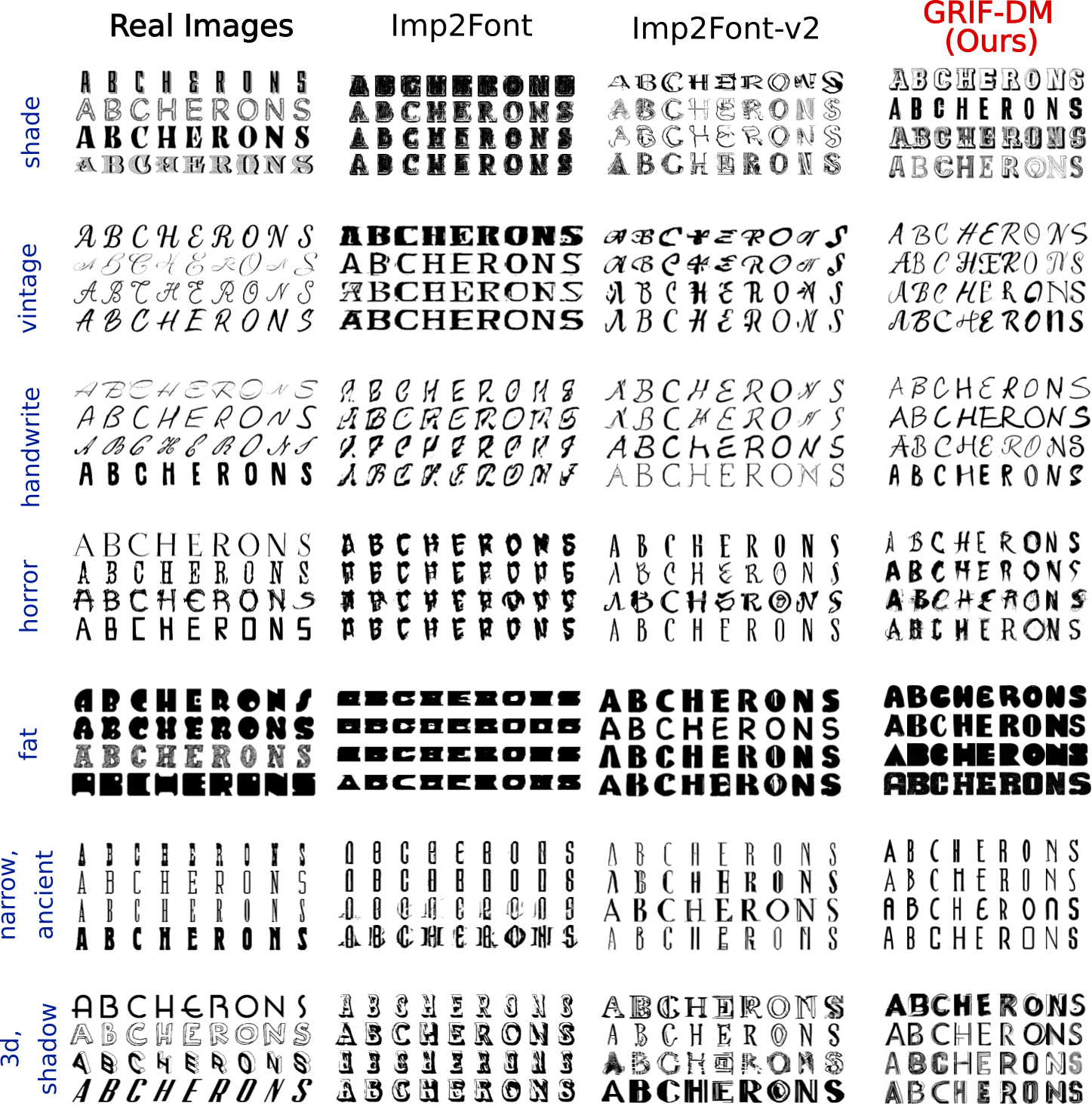}
    \caption{Font image generation with specified impression labels. Following the experimental setup from~\cite{matsuda2022font}, we utilize the letters "A", "B", "C", "H", "E", "R", "O", "N", and "S" due to their inclusion of the majority of strokes in Latin capital alphabets.}
    \label{fig:sota}
\end{figure*}

In Fig.~\ref{fig:sota}, we show the qualitative comparison with the state of the arts: Imp2Font~\cite{matsuda2021impressions2font} and Imp2Font-v2~\cite{matsuda2022font} as shown in the second and third columns, respectively. The first column depicts real font images alongside their corresponding impression keywords, shown vertically in blue. The last column showcases font images generated by our proposed method \ourmethod. The results demonstrate that our proposed method excels in both generating diverse font images and maintaining high fidelity compared to state-of-the-art methods.

In the "vintage" impression row, \ourmethod demonstrates diverse cursive strokes, exhibiting superior diversity and high-fidelity compared to other methods. In the "horror" impression row, \ourmethod impressively generates high-fidelity font images, notably with the "H" resembling axes and the "E" resembling knives. In the "fat" impression row, imp2font produces "fat" font images, but some lack readability due to textual content issues. imp2font-v2 generates thick, albeit not "fat" font images. Meanwhile, \ourmethod generates diverse "fat" font images with different round and square shapes.

In the "narrow, ancient" impression row, \ourmethod generates font images that convey the intended impression keywords but lacks the narrow fidelity of the ground truth. Similarly, state-of-the-art methods also struggle to achieve this level of narrow fidelity. In the "3d, shadow" impression row, both the state-of-the-art methods and our \ourmethod exhibit poor performance, particularly with rotated "3d" real font images, resulting in a loss of fidelity across all methods.

\vspace{0.4cm}

\subsection{Failure Cases}

In Tab.~\ref{tab:fail}, we highlight significant failure cases observed in our experiments. Due to the lack of manual filtering for the MyFont dataset, non-alphabetic symbols, such as dingbat flowers, are present, as demonstrated in the first row. Interestingly, our model is capable of generating various styles of flowers conditioned on the "flower" impression keyword. In the second row, difficulties arise in accurately generating fonts corresponding to specific impression keywords, such as "stitch", resulting in fonts resembling "gothic" instead. The issue arises due to the imbalance in keyword distribution. Lastly, while attempting to generate "funny" and "curly" fonts, our model struggles to achieve high-fidelity results, although some curvature may be discernible.

Thus, our experiments reveal challenges in accurately generating fonts corresponding to specific impression keywords, particularly when faced with imbalanced keyword distributions. Addressing these challenges could improve the performance of our model in generating diverse and faithful font images.

\begin{table}[t!]
    \caption{Failure case experiments. The fonts are "floral-orinaments", "stitch-warrior", and "kurly", from top to bottom respectively.}
    \vspace{0.4cm}
    \label{tab:fail}
    \centering
    \small
    \scalebox{0.99}{
    \begin{tabular}{ccc}
    \toprule
    \textbf{Impression Keywords} & \textbf{GT} & \textbf{Gen}\\
    \midrule
    decorative, flower, ... &
    \includegraphics[width=0.25\linewidth]{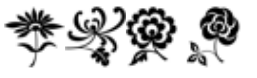} & \includegraphics[width=0.25\linewidth]{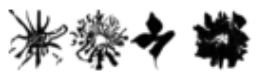}\\
    \midrule
    stitch, gothic, ... & 
    \includegraphics[width=0.25\linewidth]{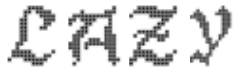} & \includegraphics[width=0.25\linewidth]{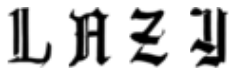}\\
    \midrule
    funny, curly, ... &
    \includegraphics[width=0.25\linewidth]{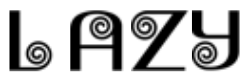} & \includegraphics[width=0.25\linewidth]{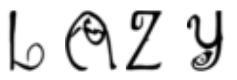}\\ 
    \bottomrule
    \end{tabular}
    }
\end{table}
\vspace{0.4cm}

\section{Conclusion and Future Work}

Our paper presents a diffusion-based method for generating fonts that are rich in impression, utilizing novel dual cross-attention modules. These modules adeptly integrate impression keywords with specific letters, facilitating a seamless generation process. Through extensive experimentation, our approach, denoted as \ourmethod, has proven effective in producing fonts that are not only realistic and vivid but also highly customized, meeting specific user demands with high fidelity.

For future work, we plan to enhance \ourmethod by incorporating Large Language Models (LLMs) into the font generation pipeline. This integration allows users to input a single natural language text, blending both the textual content and desired impression characteristics. This advancement will streamline the input process, enabling a more intuitive experience and potentially broadening the applicability of our method to a wider range of creative and commercial uses.

\minisection{Limitations.}
Our method employs a diffusion architecture for generating fonts, which, while innovative, also presents certain challenges. Firstly, the training process can be resource-intensive. Although diffusion acceleration techniques have been employed to expedite training, there remains a need for further optimization to reduce the computational overhead associated with our method. This is essential to making the approach more feasible and accessible for broader use, especially in environments with limited computational resources.
Secondly, our current focus is limited to generating fonts for the English alphabet. This limitation restricts the applicability of our method to global contexts, particularly in languages with more complex character systems, such as Chinese and Japanese. Extending our method to accommodate these and other languages presents significant challenges, not only in terms of the sheer variety of characters, but also in capturing the unique stylistic nuances each language's script entails.

\minisection{Broader Impacts.}
The adoption of personalized font generation models offers exciting prospects across a multitude of applications spanning creativity, design, and visual communication domains. Nonetheless, it is crucial to recognize potential risks associated with their deployment, such as the propagation of misinformation, potential misuse, and the introduction of biases. Ethical considerations and broader impacts necessitate a comprehensive examination to ensure the responsible utilization of these models and their capabilities.

\begin{ack}
Beatriu de Pinós del Departament de Recerca i Universitats de la Generalitat de Catalunya (2022 BP 00256), European Lighthouse on Safe and Secure AI (ELSA) from the European Union’s Horizon Europe programme under grant agreement No 101070617, Ramon y Cajal research fellowship RYC2020-030777-I / AEI / 10.13039/501100011033.
\end{ack}




\bibliography{shortbib}

\begin{thebibliography}{45}
\providecommand{\natexlab}[1]{#1}
\providecommand{\url}[1]{\texttt{#1}}
\expandafter\ifx\csname urlstyle\endcsname\relax
  \providecommand{\doi}[1]{doi: #1}\else
  \providecommand{\doi}{doi: \begingroup \urlstyle{rm}\Url}\fi

\bibitem[Amare and Manning(2012)]{amare2012seeing}
N.~Amare and A.~Manning.
\newblock Seeing typeface personality: Emotional responses to form as tone.
\newblock In \emph{IEEE International Professional Communication Conference}, pages 1--9. IEEE, 2012.

\bibitem[Azadi et~al.(2018)Azadi, Fisher, Kim, Wang, Shechtman, and Darrell]{azadi2018multi}
S.~Azadi, M.~Fisher, V.~G. Kim, Z.~Wang, E.~Shechtman, and T.~Darrell.
\newblock Multi-content gan for few-shot font style transfer.
\newblock In \emph{CVPR}, pages 7564--7573, 2018.

\bibitem[Brooks et~al.(2023)Brooks, Holynski, and Efros]{brooks2022instructpix2pix}
T.~Brooks, A.~Holynski, and A.~A. Efros.
\newblock Instructpix2pix: Learning to follow image editing instructions.
\newblock In \emph{CVPR}, 2023.

\bibitem[Chen et~al.(2022)Chen, Sun, Song, and Luo]{chen2022diffusiondet}
S.~Chen, P.~Sun, Y.~Song, and P.~Luo.
\newblock Diffusiondet: Diffusion model for object detection.
\newblock \emph{arXiv preprint arXiv:2211.09788}, 2022.

\bibitem[Chen et~al.(2019)Chen, Wang, Xu, Jin, and Luo]{chen2019large}
T.~Chen, Z.~Wang, N.~Xu, H.~Jin, and J.~Luo.
\newblock Large-scale tag-based font retrieval with generative feature learning.
\newblock In \emph{ICCV}, pages 9116--9125, 2019.

\bibitem[Devlin et~al.(2018)Devlin, Chang, Lee, and Toutanova]{bert}
J.~Devlin, M.-W. Chang, K.~Lee, and K.~Toutanova.
\newblock Bert: Pre-training of deep bidirectional transformers for language understanding.
\newblock \emph{arXiv preprint arXiv:1810.04805}, 2018.

\bibitem[Gu et~al.(2022)Gu, Chen, Xu, Lan, Meng, and Wang]{gu2022diffusioninst}
Z.~Gu, H.~Chen, Z.~Xu, J.~Lan, C.~Meng, and W.~Wang.
\newblock Diffusioninst: Diffusion model for instance segmentation.
\newblock \emph{arXiv preprint arXiv:2212.02773}, 2022.

\bibitem[Hayashi et~al.(2019)Hayashi, Abe, and Uchida]{hayashi2019glyphgan}
H.~Hayashi, K.~Abe, and S.~Uchida.
\newblock Glyphgan: Style-consistent font generation based on generative adversarial networks.
\newblock \emph{Knowledge-Based Systems}, 186:\penalty0 104927, 2019.

\bibitem[He et~al.(2022)He, Chen, Wang, Liu, Du, Tao, and Qiao]{he2022diff}
H.~He, X.~Chen, C.~Wang, J.~Liu, B.~Du, D.~Tao, and Y.~Qiao.
\newblock Diff-font: Diffusion model for robust one-shot font generation.
\newblock \emph{arXiv preprint arXiv:2212.05895}, 2022.

\bibitem[Hertz et~al.(2023)Hertz, Mokady, Tenenbaum, Aberman, Pritch, and Cohen-Or]{hertz2022prompt}
A.~Hertz, R.~Mokady, J.~Tenenbaum, K.~Aberman, Y.~Pritch, and D.~Cohen-Or.
\newblock Prompt-to-prompt image editing with cross attention control.
\newblock \emph{ICLR}, 2023.

\bibitem[Heusel et~al.(2017)Heusel, Ramsauer, Unterthiner, Nessler, and Hochreiter]{heusel2017gans}
M.~Heusel, H.~Ramsauer, T.~Unterthiner, B.~Nessler, and S.~Hochreiter.
\newblock Gans trained by a two time-scale update rule converge to a local nash equilibrium.
\newblock \emph{NeurIPS}, 30, 2017.

\bibitem[Ho et~al.(2020)Ho, Jain, and Abbeel]{ho2020denoising}
J.~Ho, A.~Jain, and P.~Abbeel.
\newblock Denoising diffusion probabilistic models.
\newblock \emph{NeurIPS}, 33:\penalty0 6840--6851, 2020.

\bibitem[Kaneko et~al.(2018)Kaneko, Ushiku, and Harada]{kaneko2018class}
T.~Kaneko, Y.~Ushiku, and T.~Harada.
\newblock Class-distinct and class-mutual image generation with gans.
\newblock \emph{arXiv preprint arXiv:1811.11163}, 2018.

\bibitem[Kang et~al.(2021)Kang, Riba, Rusinol, Fornes, and Villegas]{kang2021content}
L.~Kang, P.~Riba, M.~Rusinol, A.~Fornes, and M.~Villegas.
\newblock Content and style aware generation of text-line images for handwriting recognition.
\newblock \emph{IEEE Transactions on Pattern Analysis and Machine Intelligence}, 44\penalty0 (12):\penalty0 8846--8860, 2021.

\bibitem[Li et~al.(2023)Li, Fischer, Ke, Ding, Danelljan, and Yu]{li2023ovtrack}
S.~Li, T.~Fischer, L.~Ke, H.~Ding, M.~Danelljan, and F.~Yu.
\newblock Ovtrack: Open-vocabulary multiple object tracking.
\newblock In \emph{CVPR}, pages 5567--5577, 2023.

\bibitem[Lu et~al.(2022)Lu, Zhou, Bao, Chen, LI, and Zhu]{NEURIPS2022_260a14ac}
C.~Lu, Y.~Zhou, F.~Bao, J.~Chen, C.~LI, and J.~Zhu.
\newblock Dpm-solver: A fast ode solver for diffusion probabilistic model sampling in around 10 steps.
\newblock In S.~Koyejo, S.~Mohamed, A.~Agarwal, D.~Belgrave, K.~Cho, and A.~Oh, editors, \emph{NeurIPS}, volume~35, pages 5775--5787. Curran Associates, Inc., 2022.

\bibitem[Luo et~al.(2024)Luo, Song, Ma, Wei, Yang, and Yang]{DiffusionTrack}
R.~Luo, Z.~Song, L.~Ma, J.~Wei, W.~Yang, and M.~Yang.
\newblock Diffusiontrack: Diffusion model for multi-object tracking.
\newblock \emph{AAAI}, 2024.

\bibitem[Lv et~al.(2024)Lv, Huang, Zhang, Lin, Han, and Zeng]{lv2024diffmot}
W.~Lv, Y.~Huang, N.~Zhang, R.-S. Lin, M.~Han, and D.~Zeng.
\newblock Diffmot: A real-time diffusion-based multiple object tracker with non-linear prediction.
\newblock \emph{arXiv preprint arXiv:2403.02075}, 2024.

\bibitem[Ma et~al.(2023)Ma, Yang, Ju, Zhang, Liu, Wang, Zhang, and Wang]{ma2023diffusionseg}
C.~Ma, Y.~Yang, C.~Ju, F.~Zhang, J.~Liu, Y.~Wang, Y.~Zhang, and Y.~Wang.
\newblock Diffusionseg: Adapting diffusion towards unsupervised object discovery.
\newblock \emph{arXiv preprint arXiv:2303.09813}, 2023.

\bibitem[Matsuda et~al.(2021)Matsuda, Kimura, and Uchida]{matsuda2021impressions2font}
S.~Matsuda, A.~Kimura, and S.~Uchida.
\newblock Impressions2font: Generating fonts by specifying impressions.
\newblock In \emph{ICDAR}, pages 739--754. Springer, 2021.

\bibitem[Matsuda et~al.(2022)Matsuda, Kimura, and Uchida]{matsuda2022font}
S.~Matsuda, A.~Kimura, and S.~Uchida.
\newblock Font generation with missing impression labels.
\newblock In \emph{ICPR}, pages 1400--1406. IEEE, 2022.

\bibitem[Meng et~al.(2023)Meng, Rombach, Gao, Kingma, Ermon, Ho, and Salimans]{meng2023distillation}
C.~Meng, R.~Rombach, R.~Gao, D.~Kingma, S.~Ermon, J.~Ho, and T.~Salimans.
\newblock On distillation of guided diffusion models.
\newblock In \emph{CVPR}, pages 14297--14306, 2023.

\bibitem[Mirza and Osindero(2014)]{mirza2014conditional}
M.~Mirza and S.~Osindero.
\newblock Conditional generative adversarial nets.
\newblock \emph{arXiv preprint arXiv:1411.1784}, 2014.

\bibitem[Miyato and Koyama(2018)]{miyato2018cgans}
T.~Miyato and M.~Koyama.
\newblock cgans with projection discriminator.
\newblock In \emph{ICLR (ICLR)}, 2018.

\bibitem[Odena et~al.(2017)Odena, Olah, and Shlens]{odena2017conditional}
A.~Odena, C.~Olah, and J.~Shlens.
\newblock Conditional image synthesis with auxiliary classifier gans.
\newblock In \emph{ICML}, pages 2642--2651. PMLR, 2017.

\bibitem[Pnvr et~al.(2023)Pnvr, Singh, Ghosh, Siddiquie, and Jacobs]{pnvr2023ldznet}
K.~Pnvr, B.~Singh, P.~Ghosh, B.~Siddiquie, and D.~Jacobs.
\newblock Ld-znet: A latent diffusion approach for text-based image segmentation.
\newblock In \emph{ICCV}, pages 4157--4168, 2023.

\bibitem[Saharia et~al.(2022)Saharia, Chan, Saxena, Li, Whang, Denton, Ghasemipour, Gontijo~Lopes, Karagol~Ayan, Salimans, et~al.]{saharia2022photorealistic}
C.~Saharia, W.~Chan, S.~Saxena, L.~Li, J.~Whang, E.~L. Denton, K.~Ghasemipour, R.~Gontijo~Lopes, B.~Karagol~Ayan, T.~Salimans, et~al.
\newblock Photorealistic text-to-image diffusion models with deep language understanding.
\newblock \emph{NeurIPS}, 35:\penalty0 36479--36494, 2022.

\bibitem[Singla and Sharma(2022)]{singla2022understanding}
V.~Singla and N.~Sharma.
\newblock Understanding role of fonts in linking brand identity to brand perception.
\newblock \emph{Corporate reputation review}, 25\penalty0 (4):\penalty0 272--286, 2022.

\bibitem[Song et~al.(2020)Song, Meng, and Ermon]{song2020denoising}
J.~Song, C.~Meng, and S.~Ermon.
\newblock Denoising diffusion implicit models.
\newblock \emph{arXiv preprint arXiv:2010.02502}, 2020.

\bibitem[Song et~al.(2021)Song, Meng, and Ermon]{DDIM}
J.~Song, C.~Meng, and S.~Ermon.
\newblock Denoising diffusion implicit models.
\newblock In \emph{ICLR}, 2021.

\bibitem[Tang et~al.(2024)Tang, Wang, and van~de Weijer]{tang2023iterinv}
C.~Tang, K.~Wang, and J.~van~de Weijer.
\newblock Iterinv: Iterative inversion for pixel-level t2i models.
\newblock \emph{ICME}, 2024.

\bibitem[Tanveer et~al.(2023)Tanveer, Wang, Mahdavi-Amiri, and Zhang]{tanveer2023ds}
M.~Tanveer, Y.~Wang, A.~Mahdavi-Amiri, and H.~Zhang.
\newblock Ds-fusion: Artistic typography via discriminated and stylized diffusion.
\newblock In \emph{ICCV}, pages 374--384, 2023.

\bibitem[Tian(2017)]{tian2017zi2zi}
Y.~Tian.
\newblock zi2zi: Master chinese calligraphy with conditional adversarial networks.
\newblock \emph{Internet] https://github. com/kaonashi-tyc/zi2zi}, 3:\penalty0 2, 2017.

\bibitem[Tumanyan et~al.(2023)Tumanyan, Geyer, Bagon, and Dekel]{tumanyan2022plug}
N.~Tumanyan, M.~Geyer, S.~Bagon, and T.~Dekel.
\newblock Plug-and-play diffusion features for text-driven image-to-image translation.
\newblock \emph{CVPR}, 2023.

\bibitem[Wang et~al.(2023{\natexlab{a}})Wang, Wu, Liu, Li, Meng, and Meng]{wang2023anything}
C.~Wang, L.~Wu, X.~Liu, X.~Li, L.~Meng, and X.~Meng.
\newblock Anything to glyph: Artistic font synthesis via text-to-image diffusion model.
\newblock In \emph{SIGGRAPH Asia 2023 Conference Papers}, pages 1--11, 2023{\natexlab{a}}.

\bibitem[Wang et~al.(2023{\natexlab{b}})Wang, Yang, Yang, Butt, and van~de Weijer]{kai2023DPL}
K.~Wang, F.~Yang, S.~Yang, M.~A. Butt, and J.~van~de Weijer.
\newblock Dynamic prompt learning: Addressing cross-attention leakage for text-based image editing.
\newblock \emph{NeurIPS}, 2023{\natexlab{b}}.

\bibitem[Wang and Lian(2021)]{wang2021deepvecfont}
Y.~Wang and Z.~Lian.
\newblock Deepvecfont: synthesizing high-quality vector fonts via dual-modality learning.
\newblock \emph{ACM Transactions on Graphics (TOG)}, 40\penalty0 (6):\penalty0 1--15, 2021.

\bibitem[Wang et~al.(2020)Wang, Gao, and Lian]{wang2020attribute2font}
Y.~Wang, Y.~Gao, and Z.~Lian.
\newblock Attribute2font: Creating fonts you want from attributes.
\newblock \emph{ACM Transactions on Graphics (TOG)}, 39\penalty0 (4):\penalty0 69--1, 2020.

\bibitem[Wen et~al.(2021)Wen, Li, Han, and Yuan]{wen2021zigan}
Q.~Wen, S.~Li, B.~Han, and Y.~Yuan.
\newblock Zigan: Fine-grained chinese calligraphy font generation via a few-shot style transfer approach.
\newblock In \emph{Proceedings of the 29th ACM International Conference on Multimedia}, pages 621--629, 2021.

\bibitem[Wu et~al.(2024)Wu, Wang, Tang, and Zhang]{wu2024diffusion}
T.~Wu, K.~Wang, C.~Tang, and J.~Zhang.
\newblock Diffusion-based network for unsupervised landmark detection.
\newblock \emph{Knowledge-Based Systems}, page 111627, 2024.

\bibitem[Xie et~al.(2021)Xie, Chen, Sun, and Lu]{xie2021dg}
Y.~Xie, X.~Chen, L.~Sun, and Y.~Lu.
\newblock Dg-font: Deformable generative networks for unsupervised font generation.
\newblock In \emph{CVPR}, pages 5130--5140, 2021.

\bibitem[Xu et~al.(2023)Xu, Liu, Vahdat, Byeon, Wang, and De~Mello]{xu2023odise}
J.~Xu, S.~Liu, A.~Vahdat, W.~Byeon, X.~Wang, and S.~De~Mello.
\newblock Open-vocabulary panoptic segmentation with text-to-image diffusion models.
\newblock In \emph{CVPR}, pages 2955--2966, 2023.

\bibitem[Yang et~al.(2024)Yang, Peng, Kong, Zhang, Yao, and Jin]{yang2024fontdiffuser}
Z.~Yang, D.~Peng, Y.~Kong, Y.~Zhang, C.~Yao, and L.~Jin.
\newblock Fontdiffuser: One-shot font generation via denoising diffusion with multi-scale content aggregation and style contrastive learning.
\newblock In \emph{AAAI}, volume~38, pages 6603--6611, 2024.

\bibitem[Zhang et~al.(2023)Zhang, Rao, and Agrawala]{zhang2023adding}
L.~Zhang, A.~Rao, and M.~Agrawala.
\newblock Adding conditional control to text-to-image diffusion models.
\newblock In \emph{ICCV}, pages 3836--3847, 2023.

\bibitem[Zhang et~al.(2018)Zhang, Zhang, and Cai]{zhang2018separating}
Y.~Zhang, Y.~Zhang, and W.~Cai.
\newblock Separating style and content for generalized style transfer.
\newblock In \emph{CVPR}, pages 8447--8455, 2018.

\end{thebibliography}

\end{document}